\documentclass[preprint,12pt,authoryear]{elsarticle}

\usepackage{amssymb}

\usepackage{url} 
\usepackage{graphicx}%
\usepackage{multirow}%
\usepackage{amsmath,amssymb,amsfonts}%
\usepackage{amsthm}%
\usepackage{mathrsfs}%
\usepackage[title]{appendix}%
\usepackage{xcolor}%
\usepackage{textcomp}%
\usepackage{manyfoot}%
\usepackage{booktabs}%
\usepackage{algorithm}%
\usepackage{algorithmicx}%
\usepackage{algpseudocode}%
\usepackage{listings}%
\usepackage{placeins}

\DeclareMathAlphabet\mathbfcal{OMS}{cmsy}{b}{n}
\newcommand{\ten}[1]{\mathbfcal{#1}}
\newcommand{\mat}[1]{\mathbf{#1}}

\journal{Journal of Hydrology}

\begin{document}

\begin{frontmatter}



\title{Low-rank tensor structure of precipitation and its application to satellite–reference merging}

\author[label1,label2]{Ryan Solgi\corref{cor1}}
\ead{solgi@ucsb.edu}
\author[label3]{Rohan Shankar}
\author[label1]{Hugo A. Lo\'aiciga}
\affiliation[label1]{organization={Department of Geography, University of California-Santa Barbara}, state={CA},          country={USA}}

\affiliation[label2]{organization={Department of Electrical and Computer Engineering, University of California-Santa Barbara}, state={CA},          country={USA}}

\affiliation[label3]{organization={Department of Computer Science, Cornell University}, state={NY},          country={USA}}
\cortext[cor1]{Corresponding author.}



\begin{abstract}
The intermittent and variable nature of precipitation makes its accurate estimation over extended domains difficult, yet its spatiotemporal structure suggests that a low-rank representation may be possible. This work represents daily precipitation over the contiguous United States (CONUS) as spatiotemporal tensors and applies CANDECOMP/PARAFAC factorization, showing that preserving the native spatial and temporal modes yields more accurate reconstruction than factorizing independent daily fields or unfolded space--time matrices. Building on this finding, this work presents TMerge, a tensor-based framework that integrates satellite precipitation with sparse reference observations through shared low-rank spatial and temporal factors. TMerge was applied to correct the IMERG Final Run product with climate prediction center reference observations over CONUS. During 2019--2022, TMerge increased correlation from 0.53 to 0.85 and reduced root-mean-square error and mean absolute error by 48.2\% and 29.3\%, respectively. TMerge consistently outperformed linear bias correction, quantile mapping, and neural networks across seasons, precipitation-intensity regimes, and regions. Improvements were spatially coherent and largest in coastal regions where IMERG errors were greatest. These results demonstrate that low-rank tensor structure parsimoniously approximates the dominant spatiotemporal variability of precipitation and provides a practical mechanism for improving satellite estimates under limited reference observations over extended domains.
\end{abstract}


\begin{highlights}
\item Low-rank tensor factorization parsimoniously represents precipitation variability.
\item Tensor factorization surpasses daily and unfolded matrix-based factorizations.
\item TMerge fuses satellite precipitation and sparse references via shared tensor factors.
\item TMerge outperformed linear bias correction, quantile mapping, and neural networks.
\item Gains were largest in coastal regions, where IMERG errors were greatest.
\end{highlights}

\begin{keyword}

Low-rank precipitation \sep Tensor factorization \sep Precipitation merging \sep Precipitation variability \sep Satellite precipitation



\end{keyword}

\end{frontmatter}



\section{Introduction}

As the primary mechanism that delivers atmospheric water to the Earth's surface, precipitation underpins water resources, food production, flood and drought risks, and economic activity worldwide~\citep{Kotz2022}. Reliable precipitation estimates are therefore required for environmental assessment and decision-making~\citep{Portier2023GPM}. However, precipitation has asymmetric and intermittent spatiotemporal patterns, making its accurate estimation difficult, particularly over extended spatial and temporal domains~\citep{Sun2006, PendergrassKnutti2018,LovejoySchertzer2013}. The highly variable nature of precipitation is shaped by complex cloud microphysical and convective processes that remain challenging to represent explicitly in global models~\citep{Arakawa2004,Stevens2009,Prein2023}, and parameterized representations of these processes are associated with substantial uncertainty~\citep{Bauer2015,Yuval2020}. Therefore, direct and indirect precipitation observations, together with the gridded products derived from them, remain central to characterizing precipitation variability, evaluating models, and supporting hydrologic and environmental applications~\citep{sun2018review}.

Multi-source merging is often used to develop precipitation products by integrating complementary observations~\citep{Kidd2011Status, Sun2026PRIMER}, as ground-based measurements are relatively accurate but spatially constrained, while satellite-based estimates, with broader coverage, vary in temporal sampling and retain retrieval uncertainties~\citep{Kidd2011GPM}. Among various satellite precipitation products~\citep{Hsu1997PERSIANN, Joyce2004CMORPH}, the Integrated Multi-satellitE Retrievals for GPM (IMERG) is a leading near-global precipitation product developed within NASA’s Global Precipitation Measurement (GPM) mission and succeeding the earlier TRMM multisatellite precipitation analysis~\citep{Huffman2007TMPA}. IMERG combines passive microwave and infrared observations with monthly gauge adjustment in the Final Run, providing precipitation estimates at 0.1$^\circ$ spatial resolution~\citep{Huffman2020IMERG}. Nevertheless, errors remain substantial and heterogeneous; regional evaluations of the IMERG Final Run have reported daily root-mean-square error (RMSE) values of approximately 4.5--4.7 mm day$^{-1}$ over Europe and the contiguous United States (CONUS)~\citep{Massari2020}. These errors are particularly pronounced in coastal zones, where sharp land--ocean emissivity gradients and coastal precipitation regimes, such as orographic enhancement, make satellite precipitation detection and estimation especially difficult, as evidenced by IMERG evaluations over coastal regions of the United States~\citep{Derin2021IMERGPartI,Derin2022IMERGPartII}. These persistent errors have led to various correction and merging approaches to improve satellite precipitation products, from bias correction and distribution matching to machine learning methods using auxiliary predictors~\citep{Manz2016,Nguyen2026,Chen2021,Dao2025,Baig2025}. However, achieving consistent improvements across broad domains with diverse terrain, coastlines, and precipitation regimes remains a challenge and motivates a closer examination of the underlying structure of precipitation fields.

Low-rank methods have been applied across several hydrologic applications, including streamflow prediction~\citep{Zhang2023SparseSensing}, soil-moisture reconstruction~\citep{Beale2025GapFilling}, hydrogeologic parameter estimation~\citep{Zhuang2025LowRank}, computationally efficient inversion~\citep{Zha2018ROSLE}, data assimilation~\citep{Ghorbanidehno2015RealTime}, precipitation-monitoring network design~\citep{PrecipitationNetwork2024}, and hydroclimatic variability characterization~\citep{Zhang2021RainfallDipole}. Empirical orthogonal function analysis, a common geophysical form of principal component analysis, has also been widely used to study precipitation variability through covariance decomposition and low-rank analysis~\citep{Rao1991EOFHydrology, baldwin2009eof, Benestad2023EOF}. Precipitation exhibits a well-established spatiotemporal structure, as reflected in studies of multifractal scale invariance, stochastic rainfall modeling, and spatiotemporal modeling~\citep{veneziano2006multifractality,paschalis2013streap,banerjee2023spatial,northrop2024stochastic}. Precipitation’s inherently multidimensional spatiotemporal structure motivates moving beyond conventional low-rank analysis. Tensor factorization extends low-rank modeling to multiway data, allowing spatial, temporal, and data-source dimensions to be represented jointly~\citep{liu2022tensor}.  Tensor decomposition methods have been widely used in remote sensing for image inpainting, hyperspectral restoration, and cloud removal~\citep{Ng_weighted_tensor_2017,Zou2024tensor,zhang2024tensor}. Recently, spatiotemporal tensor approaches have also been applied to the reconstruction of environmental and geophysical data~\citep{Bai_tucker_2022, Chu_ndvi_2022,solgi2025tensorized}. In hydrology, however, tensor factorization remains underexplored particularly for precipitation estimation over extended regions, where spatial variability, temporal evolution, and multiple data sources must be modeled jointly.

This work  represents daily precipitation over extended spatiotemporal domains as tensors and applies CANDECOMP/PARAFAC (CP) factorization~\citep{Kolda2009TensorReview}, allowing spatial variability and temporal evolution to be modeled jointly while preserving the original multidimensional structure of the data. Using daily precipitation data over CONUS during 2018--2022, with the climate prediction center (CPC) unified gauge-based analysis as the reference, we show that tensor factorization reveals a relatively strong low-rank signal. This allows a compact set of low-rank factors to parsimoniously represent precipitation variability across space and time, a representation not achieved by factorizing precipitation fields as independent daily snapshots or unfolded space--time matrices. To demonstrate the practical utility of low-rank tensor structure, this study adopts a tensor-based satellite--reference merging framework, TMerge, which couples the IMERG Final Product and CPC precipitation through shared low-rank factors along spatial, temporal, and source modes. TMerge improves the IMERG Final Product over CONUS, a domain characterized by diverse hydroclimatic regimes, complex terrain, and heterogeneous precipitation processes. This work also evaluates TMerge against widely used correction baselines, including linear bias correction, neural networks, and distribution-matching methods, across regions, seasons, precipitation-intensity regimes, and varying reference-data availability. This work's results demonstrate that low-rank tensor structure provides a compact multilinear representation of precipitation variability, enabling improved satellite--reference precipitation merging over extended spatiotemporal domains under sparse reference observations.

\section{Methods}

Low-rank analysis in hydrology has remained matrix based, with spatiotemporal data commonly reshaped into space–time matrices. This unfolding does not preserve the original multidimensional structure and therefore may limit the effectiveness of low-rank representations. If tensor factorization proves effective in improving the fidelity with which the dominant variability of precipitation is captured in low-rank form it could further support hydrologic applications that rely on low-rank structure. This would produce efficient data representation, reliable reconstruction, scalable modeling, and systematic monitoring. This work first examines the hypothesis that continental-scale precipitation fields exhibit a low-rank structure that is most effectively revealed when the data are represented and factorized as spatiotemporal tensors, rather than as independent spatial snapshots or as space--time matrices. The next step presents a tensor-based merging framework that exploits shared low-rank factors between satellite precipitation products and sparse reference observations, thereby providing more accurate, high-resolution precipitation estimates.

\subsection{Tensor and low-rank representation}

This work represents precipitation fields as tensors (multidimensional arrays) to preserve their spatial, temporal, and multi-source structure. A daily gridded precipitation field can be represented as a second-order tensor (matrix), whose two modes correspond to latitude and longitude. When daily gridded precipitation fields are stacked over time, the resulting dataset forms a third-order tensor, with modes corresponding to latitude, longitude, and time. When multiple precipitation products or data sources are represented jointly, an additional source mode can be introduced, yielding a fourth-order tensor. More generally, a tensor $\ten{X} \in \mathbb{R}^{I_1 \times I_2 \times \dots \times I_N}$ is a real-valued, multidimensional, array whose entries represent precipitation estimates or observations indexed across $N$ modes. 

This work adopts CP factorization to examine and leverage the low-rank structure of spatiotemporal precipitation data. A rank-$R$ CP decomposition represents $\ten{X}$ as a sum of separable rank-one components \citep{Kolda2009TensorReview}:

\begin{align}
\label{eq:cp}
\hat{\ten{X}}(i_1,\dots,i_N)
= \sum_{r=1}^{R} \lambda_r 
\prod_{n=1}^{N} \mat{A}^{(n)}(i_n,r),
\end{align}

where $\mat{A}^{(n)} \in \mathbb{R}^{I_n \times R}$ are factor matrices along mode $n$, and $\lambda_r$ are scaling coefficients. In this paper the   term ``low rank'' denotes tensors that admit a low-rank representation, meaning that their variability can be approximated with on the order of $R \sum_{n=1}^{N} I_n$ parameters rather than $\prod_{n=1}^{N} I_n$ parameters required by their full ambient representation. Given the observed entries of $\ten{X}$, the factor matrices defining $\hat{\ten{X}}$ are obtained using alternating least squares by minimizing the mean squared error between the observed values and their corresponding estimates in $\hat{\ten{X}}$~\citep{liu2022tensor}.

\subsection{Low-Rank Spatiotemporal Tensor Modeling}
This work applies CP factorization to examine whether spatiotemporal precipitation tensors admit a parsimonious low-rank representation. This formulation is herein named low-rank spatiotemporal tensor model (STT).  STT represents daily precipitation fields within a spatial grid and time as a third-order tensor
$\ten{X}\in\mathbb{R}^{I\times J\times T}$,
where $I$ and $J$ denote the spatial dimensions and $T$ denotes time. In this case ($N=3$), the general low-rank model presented by Eq.~\eqref{eq:cp} reduces to

\begin{align}
\hat{\ten{X}}(i,j,t)
= \sum_{r=1}^{R} \lambda_r
\mat{A}(i,r)\mat{B}(j,r)\mat{C}(t,r),
\end{align}

where $\mat{A}\in \mathbb{R}^{I\times R}$ and $\mat{B} \in \mathbb{R}^{J\times R}$ describe spatial factors and $\mat{C}\in \mathbb{R}^{T \times R}$ captures temporal variability. This formulation represents precipitation variability through shared latent spatial and temporal factors, coupling all days through common spatial modes and all locations through shared temporal evolution. Therefore, the application of STT allows assessing whether or not dominant spatiotemporal components can preserve precipitation variability while reducing dimensionality.

\subsubsection{Baselines}

To distinguish the effects of temporal coupling and tensor representation, this work considers two alternative low-rank baselines: a spatial-only formulation that factorizes each daily precipitation snapshot independently and an unfolded formulation that reshapes a three-dimensional precipitation tensor into a space--time matrix. Both baselines are second-order ($N=2$) cases of Eq.~\eqref{eq:cp}; for fully observed matrices, their least-squares optimal low-rank approximations are given by the truncated singular value decomposition (SVD).


\paragraph{Spatial modeling (SM)} Precipitation in every day $t$ is modeled independently as a matrix 
$\mat{X}_t \in \mathbb{R}^{I \times J}$. 
This corresponds to the $N=2$ case of the low-rank tensor model presented by Eq.~\eqref{eq:cp} and the low-rank decomposition reduces to

\begin{align}
\hat{\mat{X}}_t(i,j)
= \sum_{r=1}^{R'} \lambda_{t,r}
\mat{A}_t(i,r)\mat{B}_t(j,r),
\end{align}

where $\mat{A}_t \in \mathbb{R}^{I \times R'}$, and $\mat{B}_t \in \mathbb{R}^{J \times R'}$ denote the spatial factors at day $t$, with $R'$ the prescribed rank. In contrast to the fully coupled spatiotemporal model, this baseline does not share information across time. 

\paragraph{Unfolded modeling (UM)} The spatiotemporal tensor $\ten{X} \in \mathbb{R}^{I \times J \times T}$ is unfolded into a matrix $\mat{X} \in \mathbb{R}^{IJ \times T}$ by flattening the two spatial modes into a single spatial dimension, such that each row corresponds uniquely to one latitude--longitude grid cell. The resulting matrix is factorized as

\begin{align}
\hat{\mat{X}}(s,t)
= \sum_{r=1}^{R''} \lambda_{r}
\mat{U}(s,r)\mat{V}(t,r),
\end{align}

where $\mat{U} \in \mathbb{R}^{IJ \times R''}$ and $\mat{V} \in \mathbb{R}^{T \times R''}$ denote the spatial and temporal factors, respectively, with $R''$ the prescribed rank. Thus, UM retains joint space--time coupling but loses the original two-dimensional spatial structure preserved by STT.

\subsection{Satellite--reference precipitation merging}

To leverage the low-rank tensor structure of precipitation, this work introduces a tensor-based merging framework, referred to as TMerge, for improving satellite precipitation estimates using sparse reference observations. The rationale behind TMerge is that the dominant latent low-rank structure underlying the spatiotemporal variability of precipitation should be at least partly shared between satellite and reference products, even though the two sources differ in accuracy, coverage, and error structure. TMerge exploits this latent structure by factorizing a joint satellite--reference precipitation tensor into shared low-rank spatial and temporal factors. In this formulation, the satellite product provides high resolution spatial coverage of precipitation variability, while the sparse reference observations provide more reliable constraints. By combining these complementary strengths, the joint low-rank representation produces a high-resolution precipitation product corrected by the reference observations.

\subsubsection{TMerge: Low-rank tensor merging}

TMerge extends the spatiotemporal low-rank formulation from a single precipitation tensor to a joint satellite--reference representation. This joint tensor is constructed by introducing an additional source mode, yielding
$\ten{Y} \in \mathbb{R}^{I \times J \times T \times L}$, where $L=2$ corresponds to the satellite ($\mathrm{sat}$) and reference ($\mathrm{ref}$) sources. This corresponds to the $N=4$ case of the low-rank model:

\begin{align}
\hat{\ten{Y}}(i,j,t,l)
= \sum_{r=1}^{R} \lambda_r
\mat{A}(i,r)\mat{B}(j,r)\mat{C}(t,r)\mat{D}(l,r),
\end{align}

where $\mat{A}\in\mathbb{R}^{I\times R}$ and $\mat{B}\in\mathbb{R}^{J\times R}$ represent spatial factors, $\mat{C}\in\mathbb{R}^{T\times R}$ represents temporal factors, and $\mat{D}\in\mathbb{R}^{L\times R}$ captures source-specific loadings. The satellite source is observed over the full spatial grid, whereas the reference source is observed only at the reference locations. Both sources share the same spatial and temporal factor matrices within the low-rank formulation. The satellite entries provide dense spatiotemporal information, while the sparse reference entries constrain the same latent factors toward the reference field. By imposing a multilinear structure across space, time, and source, TMerge then estimates the precipitation value at an unseen location $(i,j)$ at time $t$ as $\hat{X}_{i,j,t}=\hat{\ten{Y}}(i,j,t,\mathrm{ref})$.

\subsubsection{Baseline correction methods}
This work compares TMerge with three satellite-precipitation correction baselines: multiplicative linear correction (Linear), a coordinate-informed multilayer perceptron neural network (MLP), and quantile mapping with spatial interpolation (QM). These baselines cover a range of correction strategies, including global linear scaling, distribution matching, and nonlinear modeling providing a comprehensive evaluation of TMerge.

\paragraph{Linear correction} Multiplicative linear bias-correction is a widely used approach to reduce systematic biases in satellite precipitation products \citep{Manz2016}. A linear correction estimates the precipitation value at an unseen location $(i,j)$ at time $t$ by scaling the satellite value as 

\begin{align}
\hat{X}_{i,j,t}=\alpha X_{i,j,t}^{\mathrm{sat}},
\end{align}

where the coefficient $\alpha$ is estimated using paired satellite and reference observations at available locations by minimizing mean squared errors.

\paragraph{Quantile mapping}
Quantile mapping (QM) corrects satellite estimates by matching their distribution to
that of the reference observations \citep{Nguyen2026}. At each reference location
$s_r$, an empirical map $\hat{h}_{s_r}(x) = \hat{F}^{-1}_{s_r}(\hat{G}_{s_r}(x))$
is fitted where $\hat{F}_{s_r}$ and
$\hat{G}_{s_r}$ are the empirical cumulative distribution functions of the
satellite and reference precipitation at $s_r$. The reference corrections 
$e_{s_r,t} = \hat{h}_{s_r}(X^{\mathrm{sat}}_{s_r,t}) - X^{\mathrm{sat}}_{s_r,t}$ for all $s_r$ at time $t$ are interpolated to have:
\begin{align}
\hat{X}_{i,j,t} = X^{\mathrm{sat}}(i,j,t)
+ \hat{e}_{i,j,t},
\end{align}
where $\hat{e}(i,j,t)$ denotes the interpolated correction at unseen gridded location $(i,j)$ and time $(t)$. An inverse distance weighting (IDW) is used for interpolation and negative predictions are set to zero.

\paragraph{Neural-network}
Deep learning and neural networks have been used for precipitation correction and prediction~\citep{Dao2025,Baig2025}. Accordingly, this work implements an MLP baseline for satellite--reference merging that estimates precipitation from the satellite value together with normalized spatial and temporal coordinates. We define these coordinates as $\tilde{i}=i/(I-1)$, $\tilde{j}=j/(J-1)$, and $\tilde{t}=t/(T-1)$. Let $f_{\theta}$ denote a network fitted for a temporal window of length $T$. The precipitation value at spatial location $(i,j)$ at time $t$ is estimated as
\begin{align}
\hat{{X}}_{i,j,t} = f_{\theta}\left({X}^{\mathrm{sat}}_{i,j,t}\, , \tilde{i}, \tilde{j}, \tilde{t}\right).
\end{align}
By using both spatial position and time as inputs, the MLP provides a relevant nonlinear correction baseline for assessing whether TMerge offers benefits beyond coordinate-informed error correction. The MLP baseline consists of fully connected hidden layers with ReLU activations~\citep{Nair2010ReLU} and a linear output layer. The network parameters are trained on the observed satellite--reference pairs using a mean squared error loss and the AdamW optimizer~\citep{Loshchilov2019AdamW}. Inputs and targets are standardized during training and transformed back to mm\,day$^{-1}$ for prediction.

\subsection{Evaluation metrics}

Model performance was evaluated across the experiments using complementary metrics as relevant to each experiment, including the Pearson correlation coefficient ($r$), which measures the linear association between estimated and observed precipitation~\citep{legates1999}; the root-mean-square error (RMSE), which measures error magnitude while placing greater emphasis on large deviations~\citep{chai2014}; the mean absolute error (MAE), which measures the average absolute difference between estimated and observed precipitation values~\citep{willmott2005}; the Nash--Sutcliffe efficiency (NSE), which evaluates predictive performance relative to the mean of the observations~\citep{nash1970}; and the Kling--Gupta efficiency (KGE), which jointly accounts for correlation, bias, and variability~\citep{gupta2009}.

\section{Results}
This work organizes results according to two questions: whether daily precipitation fields admit a compact spatiotemporal low-rank representation, and whether that structure can be exploited to merge sparse gauge-based references with satellite estimates. To address the first question,  STT is applied to CPC daily precipitation fields within CONUS and demonstrate that tensor factorization yields a more parsimonious low-rank representation than matrix-based formulations. To address the second question,  TMerge is applied to merge sparse CPC references with the IMERG Final daily product within CONUS. Across all experiments, fixed-location sampling is applied such that the same subset of spatial locations is used for model fitting across all time steps, while the remaining locations are reserved for evaluation. This choice reflects the fixed spatial coverage of ground-based reference networks such as rain-gauge stations.

\subsection{Low-rank approximation of precipitation tensors}

\begin{table}[!t]
\caption{Performance comparison of STT versus SM and UM baselines for daily precipitation reconstruction across CONUS during 2018--2022 under a 10\% withholding ratio. Values are mean $\pm$ maximum deviation across 5 random runs with different withholding sets. Parameter budget denotes the percentage of model parameters relative to the total number of data entries. The best method per metric is shown in bold.}
\centering
\renewcommand{\arraystretch}{1.1}
\setlength{\tabcolsep}{4pt}
\small
\begin{tabular*}{\textwidth}{@{\extracolsep{\fill}}llcc}
\toprule
Parameter Budget & Method & Correlation Coefficient ($r$) & RMSE \\
\midrule
\multirow{3}{*}{0.5\%} & STT & \textbf{0.791 $\pm$ 0.001} & \textbf{3.926 $\pm$ 0.009} \\
 & SM & 0.734 $\pm$ 0.045 & 4.413 $\pm$ 0.399 \\
 & UM & 0.343 $\pm$ 0.001 & 6.053 $\pm$ 0.032 \\
\midrule
\multirow{3}{*}{1.0\%} & STT & \textbf{0.845 $\pm$ 0.001} & \textbf{3.435 $\pm$ 0.013} \\
 & SM & 0.668 $\pm$ 0.223 & 5.266 $\pm$ 3.028 \\
 & UM & 0.343 $\pm$ 0.002 & 6.054 $\pm$ 0.012 \\
\midrule
\multirow{3}{*}{2.0\%} & STT & \textbf{0.885 $\pm$ 0.001} & \textbf{2.987 $\pm$ 0.008} \\
 & SM & 0.738 $\pm$ 0.012 & 4.383 $\pm$ 0.124 \\
 & UM & 0.436 $\pm$ 0.002 & 5.813 $\pm$ 0.021 \\
\midrule
\multirow{3}{*}{5.0\%} & STT & \textbf{0.926 $\pm$ 0.001} & \textbf{2.431 $\pm$ 0.009} \\
 & SM & 0.786 $\pm$ 0.085 & 4.321 $\pm$ 1.165 \\
 & UM & 0.526 $\pm$ 0.001 & 5.526 $\pm$ 0.013 \\
\midrule
\multirow{3}{*}{10.0\%} & STT & \textbf{0.948 $\pm$ 0.000} & \textbf{2.035 $\pm$ 0.002} \\
 & SM & 0.855 $\pm$ 0.014 & 3.570 $\pm$ 0.193 \\
 & UM & 0.607 $\pm$ 0.001 & 5.263 $\pm$ 0.018 \\
\midrule
\multirow{3}{*}{20.0\%} & STT & \textbf{0.962 $\pm$ 0.000} & \textbf{1.752 $\pm$ 0.010} \\
 & SM & 0.816 $\pm$ 0.006 & 4.119 $\pm$ 0.103 \\
 & UM & 0.629 $\pm$ 0.001 & 5.473 $\pm$ 0.021 \\
\bottomrule
\end{tabular*}
\label{tab:completion_results}
\end{table}

This section examines whether daily precipitation fields admit a compact spatiotemporal low-rank representation. For every 3-month period during 2018--2022, CPC daily precipitation data over CONUS are represented as a three-dimensional space--time tensor, and STT is fitted to every period.

\begin{figure}[t]
    \centering
    \includegraphics[width=\linewidth]{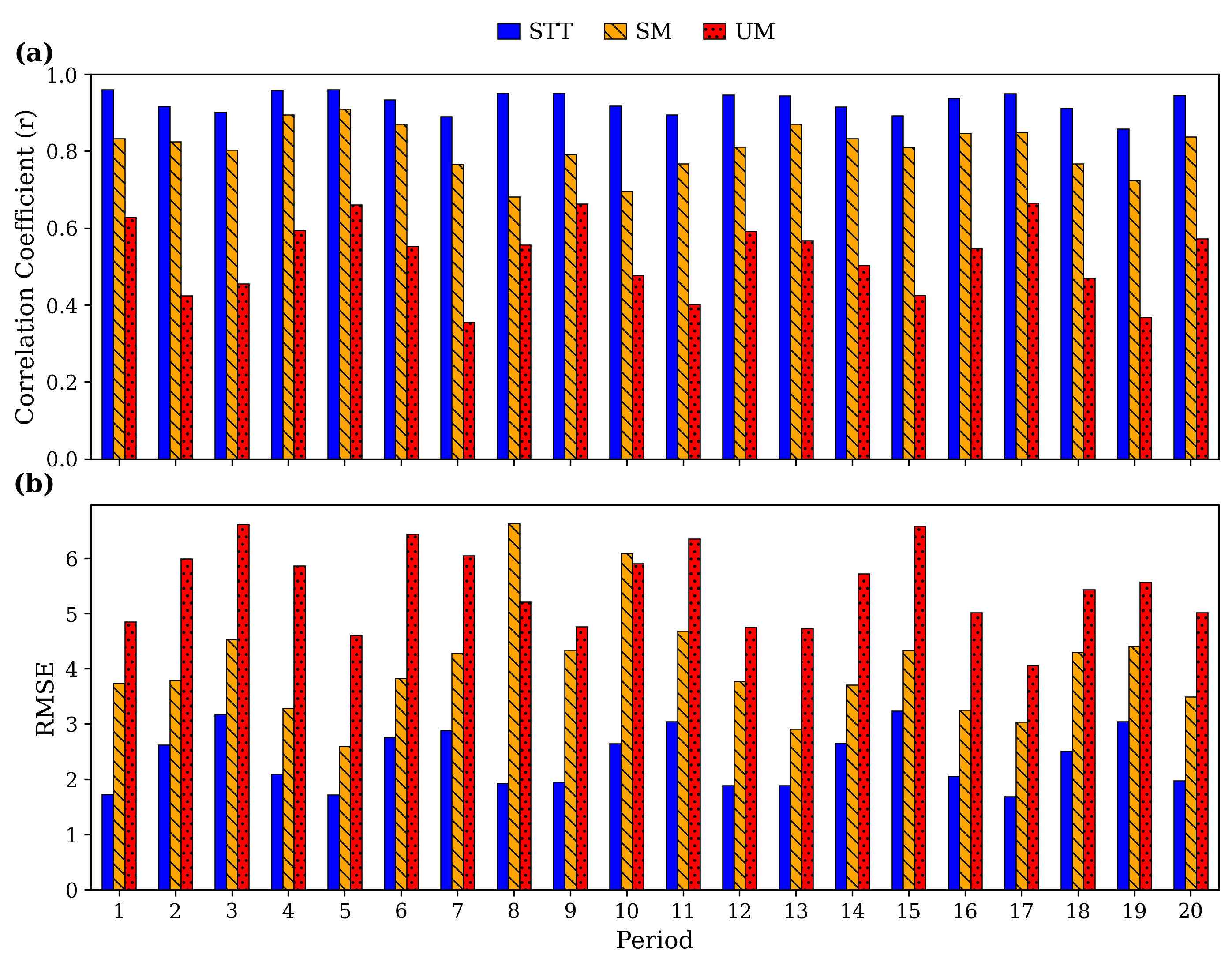}
    \caption{(a) Correlation coefficient ($r$) and (b) RMSE of STT, SM, and UM against the CPC reference over CONUS across 20 non-overlapping three-month periods during 2018--2022, using a 10\% withholding ratio and a matched parameter budget of about 5.0\%. Bars show averages over five runs.}
\label{fig:seasonal_completion}
\end{figure}

Table~\ref{tab:completion_results} compares the reconstruction accuracy of STT with SM, and UM, across the full 2018--2022 study period corresponding to a 10\% withholding ratio. To assess how much of the precipitation field can be recovered from a compact low-rank spatiotemporal representation, the rank is varied to obtain different parameter budgets; here, the parameter budget denotes the number of model parameters expressed as a percentage of the total number of entries in the full space--time precipitation tensor.  STT consistently outperforms the spatial-only and unfolded matrix factorization baselines under matched parameter budgets, achieving the highest correlation coefficient and the lowest RMSE. To account for the effect of spatial sampling, each experiment was repeated five times with a different random assignment of spatial locations to the fitting and evaluation sets. At a parameter budget of 5\%, STT achieved a correlation coefficient of \(0.926 \pm 0.001\) and RMSE of \(2.431 \pm 0.009\), outperforming both baselines on accuracy and stability. UM was the least accurate (\(r = 0.526 \pm 0.001\), RMSE \(= 5.526 \pm 0.013\)) but highly stable across runs, while SM was more accurate than UM (\(r = 0.786 \pm 0.085\), RMSE \(= 4.321 \pm 1.165\)) yet far more sensitive to the spatial split, as reflected in its much larger variability. STT combined the best of SM and UM, improving correlation by \(17.8\%\) and \(76.0\%\) over SM and UM, respectively, and reducing RMSE by \(43.7\%\) and \(56.0\%\), while retaining a tight run-to-run variance. The larger variation observed for SM reflects its greater sensitivity to the spatial split, because it needs separate factors for each daily field, whereas STT and UM estimate shared factors using observations across the full time period. The STT maintained its superiority across the full range of parameter budgets tested, from 0.5\% to 20\%, indicating that it is robust to changes in the parameter budget. The high reconstruction accuracy achieved by STT at small parameter ratios, together with its improvement over low-rank baselines suggests that preserving the native spatiotemporal structure as a multi-dimensional tensor enables a more parsimonious low-rank representation of daily precipitation fields. In contrast, treating the data as independent spatial snapshots or unfolding the space--time field into a matrix weakens the effectiveness of low-rank modeling. These results indicate that spatiotemporal tensor factorization can exploit a compact low-rank structure in precipitation fields that is not captured as effectively by spatial-only or unfolded space--time matrix representations.

\begin{figure}[t]
    \centering
    \includegraphics[width=0.6\linewidth]{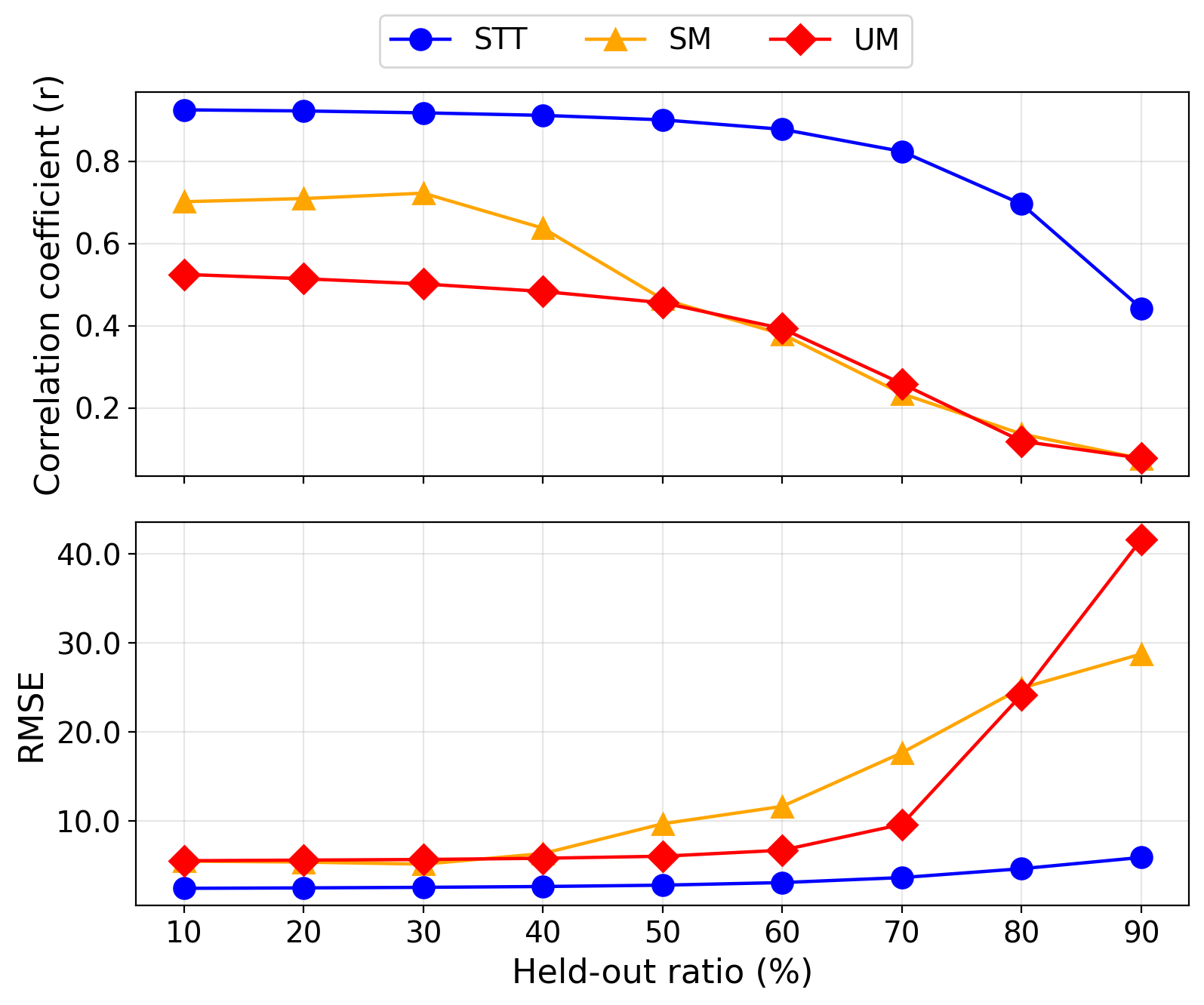}
    \caption{Comparison of low-rank models at a 5\% parameter budget under different withholding ratios.}
\label{fig:completion_result}
\end{figure}

Figure~\ref{fig:seasonal_completion} compares the correlation coefficient and RMSE of STT, SM, and UM across all 20 three-month periods during 2018--2022 at a matched parameter budget of approximately 5\%. STT achieved the highest correlation coefficient and the lowest RMSE in every period. The separation between STT and UM was especially pronounced: the lowest STT correlation across all periods ($r=0.858$) remained higher than the highest UM correlation ($r=0.665$), and the largest STT RMSE (3.23~mm~day$^{-1}$) remained lower than the smallest UM RMSE (4.05~mm~day$^{-1}$). Overall, the superiority of STT was maintained across three-month intervals spanning relatively wetter and drier portions of the annual precipitation cycle throughout the five-year study period.

Figure~\ref{fig:completion_result} shows the robustness of the low-rank spatiotemporal tensor representation under reduced spatial observation coverage, evaluated by varying the withholding ratio and comparing STT with other low-rank baselines, including SM and UM. For example, a 70\% withholding ratio means that 30\% of the CPC reference locations are used for training, while the remaining 70\% are withheld for evaluation. As spatial training coverage decreased, STT maintained substantially higher reconstruction accuracy than both baselines. In particular, STT preserved high correlation with a 70\% withholding ratio ($r=0.824$) and remained reasonably accurate at 80\% ($r=0.696$), whereas SM and UM deteriorated much more rapidly, with correlations decreasing to $r=0.235$ and $r=0.259$ at 70\%, and to $r=0.137$ and $r=0.120$ at 80\%, respectively. The RMSE results show the same behavior. STT remained below 4 mm day$^{-1}$ with the 70\% withholding ratio, increasing from 2.434 mm day$^{-1}$ at 10\% to 3.640 mm day$^{-1}$ at 70\%, and remained below 5 mm day$^{-1}$ at 80\% with an RMSE of 4.636 mm day$^{-1}$. In contrast, SM began to fail at moderate sparsity, with RMSE increasing to 9.680 mm day$^{-1}$ at 50\%, 17.656 mm day$^{-1}$ at 70\%, and 24.921 mm day$^{-1}$ at 80\%. UM was more stable than SM at moderate withholding ratios but also degraded sharply under severe sparsity, increasing from 6.713 mm day$^{-1}$ at 60\% to 9.617 mm day$^{-1}$ at 70\% and 24.160 mm day$^{-1}$ at 80\%. This indicates that the spatiotemporal tensor structure remains informative even under substantially sparse reference coverage, although the reconstruction begins to degrade at the most extreme 90\% withholding ratio (10\% reference availability).

This paper's results show that continental-scale precipitation fields contain a highly exploitable parsimonious low-rank structure, which is most effectively revealed when the data are represented and factorized as spatiotemporal tensors. The observed spatiotemporal relative low-rankness is not merely a compression artifact, but a meaningful organizing property of precipitation variability with important implications for hydrology, scalable modeling, and Earth system science.

\subsection{Low-rank tensor merging results}

To leverage the low-rank tensor structure of precipitation,  TMerge was applied for improving satellite precipitation estimates using sparse reference observations.  TMerge was evaluated in two complementary settings. First, to isolate the effect of low-rank spatiotemporal fusion under matched spatial resolution, IMERG was regridded to the CPC grid, while CPC observations were retained only at sparse fixed locations and withheld elsewhere for evaluation. This controlled setting tests whether sparse reference constraints can improve a dense satellite field when both sources are represented on the same grid. Second,  TMerge was applied on the native IMERG grid, where sparse CPC observations are used as reference constraints to generate a precipitation product at the higher IMERG spatial resolution. In both settings,  TMerge was evaluated against the uncorrected IMERG Final product and common correction baselines, including linear bias correction, a multilayer perceptron (MLP), and quantile mapping.

\subsubsection{CPC resolution experiment}

\begin{figure}[]
\centering
\includegraphics[width=0.65\textwidth]{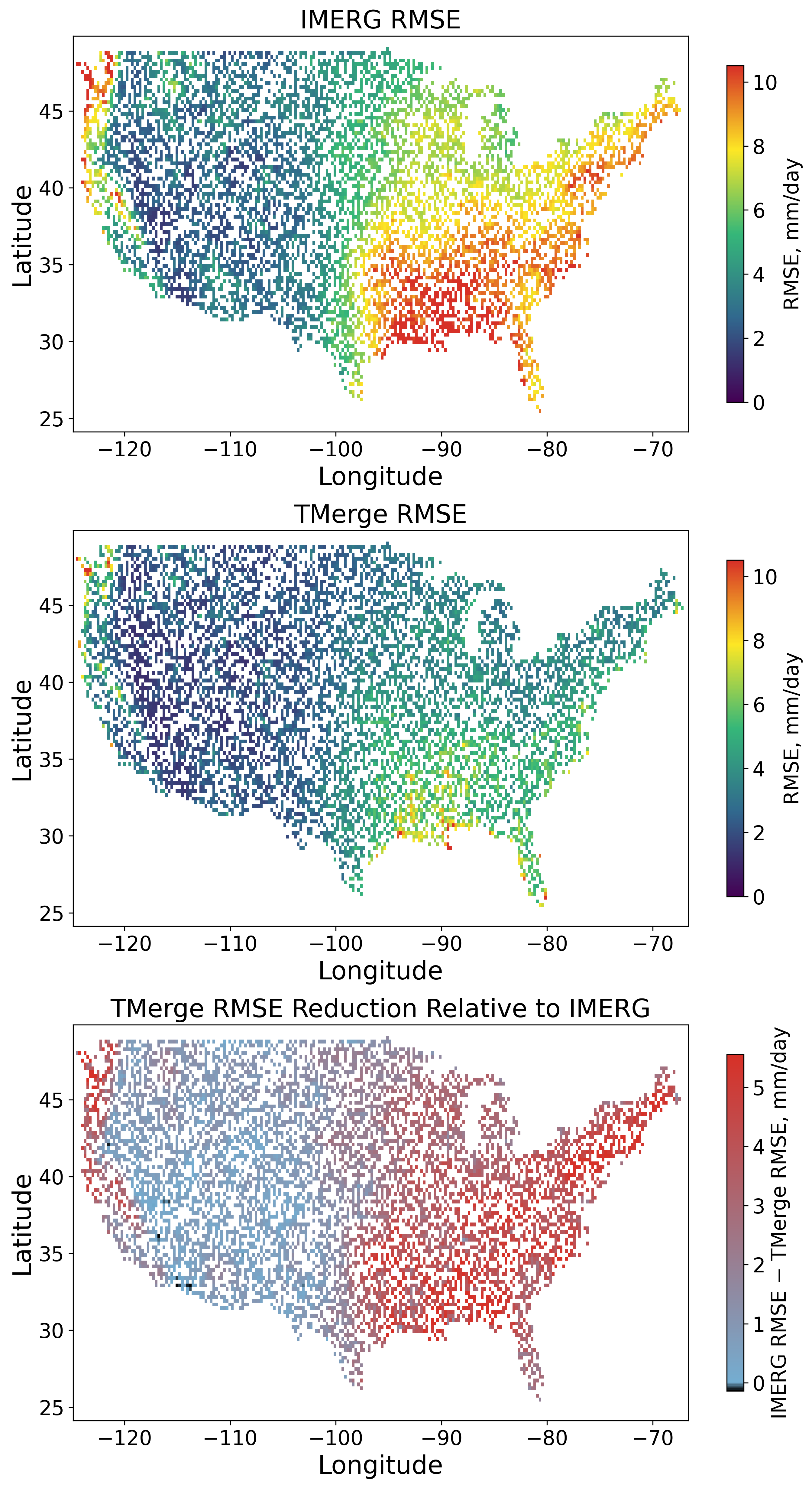}
\caption{Spatial distribution of RMSE for IMERG and TMerge daily precipitation estimates, together with the RMSE reduction of TMerge relative to IMERG, over CONUS during 2018--2022. White cells inside CONUS denote fitting locations excluded from evaluation, while surrounding white areas outside the CONUS region are not included in either model fitting or evaluation.}
\label{fig:rmse_maps}
\end{figure}

\begin{table}
\caption{Performance comparison of TMerge, MLP, Linear, QM, and IMERG for daily precipitation estimation across CONUS during 2018--2022. Values are reported as mean $\pm$ standard deviation across five random splits of CPC reference and evaluation locations. The best method per metric is shown in bold.}
\centering
\renewcommand{\arraystretch}{1.25}
\setlength{\tabcolsep}{4pt}
\begin{tabular*}{\textwidth}{@{\extracolsep{\fill}}lccc}
\toprule
Method & Correlation Coefficient ($r$) & RMSE & MAE \\
\midrule
TMerge & \textbf{0.819 $\pm$ 0.001} & \textbf{3.716 $\pm$ 0.012} & \textbf{1.903 $\pm$ 0.006} \\
MLP & 0.577 $\pm$ 0.002 & 5.205 $\pm$ 0.009 & 2.278 $\pm$ 0.014 \\
Linear & 0.542 $\pm$ 0.001 & 6.085 $\pm$ 0.008 & 2.251 $\pm$ 0.004 \\
QM & 0.564 $\pm$ 0.001 & 5.879 $\pm$ 0.008 & 2.202 $\pm$ 0.004 \\
IMERG & 0.542 $\pm$ 0.001 & 6.296 $\pm$ 0.007 & 2.329 $\pm$ 0.004 \\
\bottomrule
\end{tabular*}
\label{tab:total_method_comparison}
\end{table}

As reported in Table~\ref{tab:total_method_comparison}, TMerge substantially improved daily precipitation estimates across CONUS during 2018--2022 compared to the IMERG Final product and correction baselines. On average, the IMERG Final product achieved \(r=0.542\), RMSE \(=6.296\) mm day\(^{-1}\), and MAE \(=2.329\) mm day\(^{-1}\). The Linear correction baseline provided only modest improvements, reducing RMSE by \(3.4\%\) and MAE by \(3.3\%\). QM further improved performance, reducing RMSE by \(6.6\%\) and MAE by \(5.5\%\), while the MLP baseline reduced RMSE by \(17.3\%\) and MAE by \(2.2\%\). In contrast, TMerge improved the correlation coefficient to \(0.819\), while reducing RMSE by \(41.0\%\) and MAE by \(18.3\%\), reaching RMSE \(=3.716\) mm day\(^{-1}\) and MAE \(=1.903\) mm day\(^{-1}\). The small standard deviations across five random reference--evaluation splits indicate that the improvement is robust to the spatial sampling of CPC reference observations. Overall, these results show that TMerge provides a substantial improvement over both the uncorrected IMERG Final product and correction baselines, demonstrating that low-rank spatiotemporal fusion captures information that is not fully exploited by linear bias correction, distribution-matching, or pointwise nonlinear correction.

\begin{figure}[]
    \centering
\includegraphics[width=\linewidth]{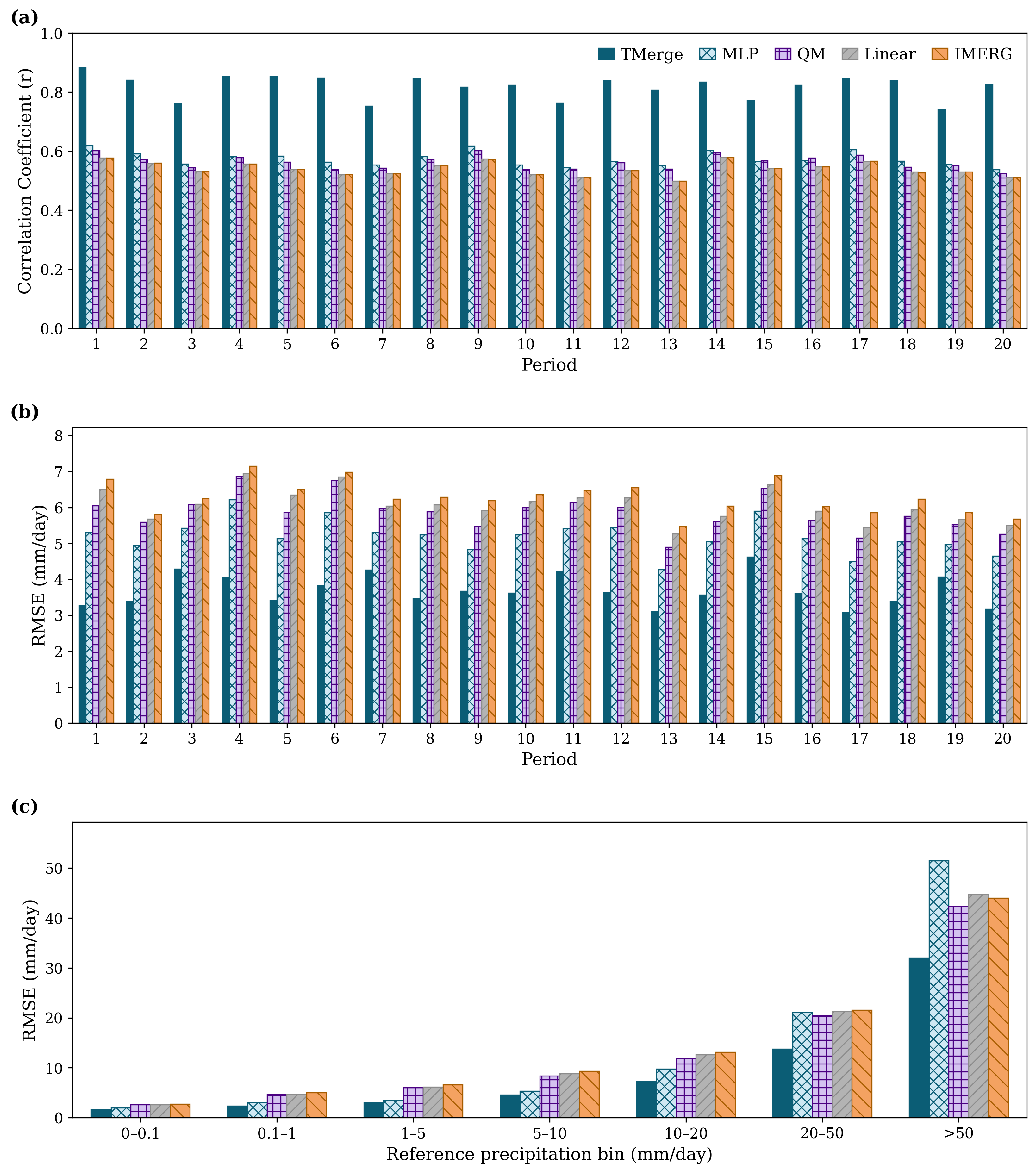}
    \caption{Performance comparison of TMerge, MLP, QM, Linear correction, and the IMERG Final product against the CPC reference at CPC resolution over CONUS during 2018--2022. (a) Correlation coefficient ($r$) across three-month periods. (b) RMSE across three-month periods. (c) RMSE across CPC reference precipitation bins.}
    \label{fig:seasonal_bins_merge}
\end{figure}

As seen in Figure~\ref{fig:rmse_maps}, the improvement was spatially coherent across CONUS. TMerge reduced grid-cell RMSE over time by an average of 2.291 mm day\(^{-1}\) compared to IMERG, with local RMSE differences ranging from a maximum degradation of 0.138 mm day\(^{-1}\) to a maximum improvement of 7.312 mm day\(^{-1}\). Nearly all evaluated grid cells (99.74\%) showed an improvement in RMSE, while minor increases in RMSE were limited to 0.26\% of the domain. Thus, the few locations with degraded performance were spatially negligible and their error increases were small compared to the average and maximum RMSE reductions across the domain. Satellite precipitation errors are known to be spatially heterogeneous. In our experiment, this heterogeneity manifested as elevated IMERG RMSE, particularly at three primary marine coastlines of the continental United States: the West Coast, the Gulf Coast, and the East Coast. These coastal zones span diverse precipitation regimes, including orographic precipitation along the mountainous West Coast, tropical and convective systems along the Gulf Coast, and complex coastal and frontal processes along the East Coast~\citep{Derin2022IMERGPartII}. Precipitation estimation in these coastal zones is challenging for satellite retrievals such as IMERG because the underlying precipitation processes in these regions are governed by terrain forcing, land--ocean gradients, and mesoscale storm dynamics that are not fully resolved by remotely sensed cloud and ice signatures. TMerge consistently improved precipitation estimation across CONUS, with the strongest error reductions in coastal regions where IMERG errors were highest. This spatial correspondence suggests that regional-scale low-rank spatiotemporal structure provides a basis for mitigating systematic components of IMERG error, and TMerge exploits this structure by propagating sparse CPC reference information through a joint tensor factorization of CPC and IMERG, thereby substantially reducing IMERG error.

The improvement was consistent across both time and precipitation regimes. As shown in Figures~\ref{fig:seasonal_bins_merge}a and~\ref{fig:seasonal_bins_merge}b across the 20 three-month periods from 2018 to 2022, TMerge achieved the highest correlations and lowest RMSE in every period, outperforming MLP, QM, Linear correction, and the IMERG Final product. Compared to IMERG, TMerge improved RMSE in all periods, with improvements ranging from 1.798 to 3.520 mm/day and a mean improvement of 2.595 mm/day. These gains were substantially larger than those of the baselines: the smallest RMSE improvement achieved by TMerge (1.798 mm/day) exceeded the largest improvements obtained by MLP (1.480 mm/day), QM (0.740 mm/day), and Linear correction (0.404 mm/day). On average across all periods, TMerge increased the correlation by a factor of 1.52 relative to IMERG, while IMERG RMSE remained 1.70 times larger than TMerge RMSE; these average gains were well above those achieved by MLP, QM, and Linear correction. The improvement was also consistent across precipitation regimes. As shown in Figure~\ref{fig:seasonal_bins_merge}c, TMerge achieved lower RMSE than IMERG in every precipitation bin, outperforming the MLP, QM, and Linear correction baselines in all regimes. TMerge RMSE improvement relative to IMERG ranged from 27\% in the \textgreater{}50 mm/day bin to 54\% in the 1--5 mm/day bin, with improvements of about 39--54\% across the 0--20 mm/day bins and a still substantial 36\% improvement in the 20--50 mm/day bin. This consistency is important because higher precipitation intensities are both less frequent and associated with larger IMERG errors, making these conditions progressively more difficult to correct. The robustness of TMerge was most evident in the \(>50\) mm/day bin. The \(>50\) mm/day bin represents a rare high-precipitation regime, accounting for only 0.28\% of all samples. In this regime, QM reduced RMSE by only about 4\%, while MLP and Linear correction degraded IMERG performance; in contrast, TMerge reduced RMSE by 27\%. These results indicate that TMerge improves the dominant precipitation regimes while remaining effective for rare high-intensity events, where IMERG errors are largest and correction is most difficult.

\begin{figure}[t]
    \centering
    \includegraphics[width=\linewidth]{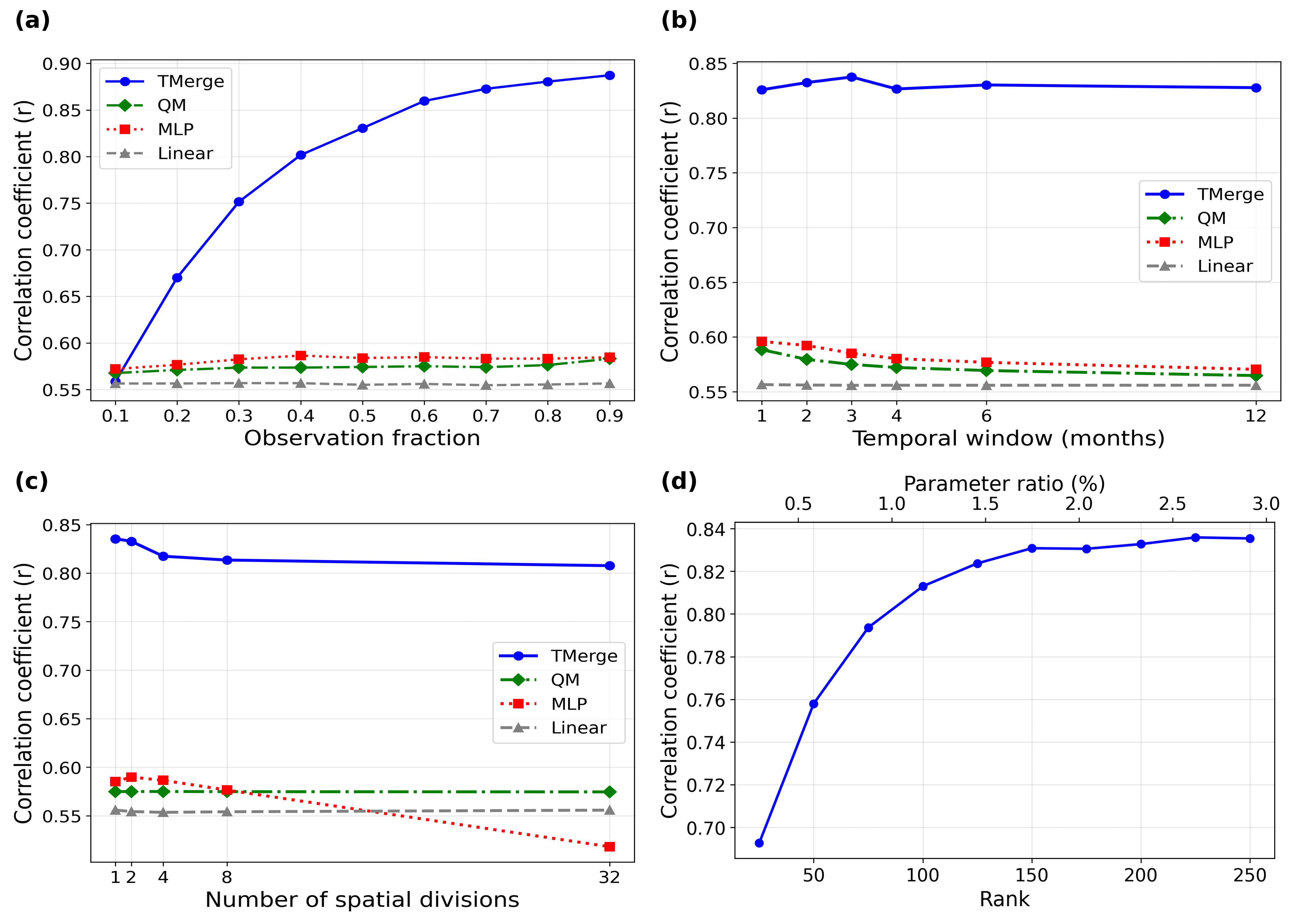}

    \caption{Sensitivity analysis of TMerge performance over CONUS in 2018 compared to baseline methods. Correlation coefficient ($r$) is shown as a function of (a) reference observation fraction, (b) temporal window length, (c) number of spatial divisions, and (d) CP rank and corresponding parameter ratio.}
    \label{fig:sens_anal}
\end{figure}

The main results discussed above were obtained using 50\% of the CPC reference locations as observations, with the remaining 50\% held out for evaluation, providing a conservative assessment of TMerge rather than its saturated performance. As shown in Figure~\ref{fig:sens_anal}a, TMerge continued to improve as additional CPC reference locations became available in 2018, with the correlation coefficient increasing from \(r=0.830\) at the 50\% observation ratio to \(r=0.887\) when 90\% of CPC locations were used for fitting. At the same time, TMerge remained competitive under sparse reference coverage: with only 10\% observation ratio, TMerge achieved \(r=0.559\), comparable to QM (\(r=0.568\)), MLP (\(r=0.572\)), and Linear correction (\(r=0.557\)); with 20\% observation ratio, TMerge increased to \(r=0.670\), already exceeding all baselines, which remained near \(r\approx0.56\text{--}0.58\). Unlike TMerge, the baselines including QM, MLP, and Linear showed limited gains from additional reference observations along the full range of observation fractions. This behavior makes TMerge relevant both for data-limited regions and for observation-rich settings where denser reference coverage can further improve precipitation estimation.

Figure~\ref{fig:sens_anal} further summarizes the sensitivity of TMerge to temporal partitioning, spatial partitioning, and CP rank for the year 2018. Figure~\ref{fig:sens_anal}b evaluates the sensitivity of TMerge to temporal partitioning while keeping the spatial partitioning fixed. TMerge remains nearly stable across partitioning choices from 1-month to 12-month windows, with correlation varying only from \(r=0.826\) to \(r=0.837\). The best performance is obtained with the 3-month window, which is used in the main results discussed above. Figure~\ref{fig:sens_anal}c evaluates the effect of spatial partitioning while the temporal window is fixed at three months. TMerge achieved its best performance when the domain was modeled as a single CONUS-wide tensor, with \(r=0.835\), and performance decreased modestly as the domain was divided into smaller spatial tiles, reaching \(r=0.808\) for 32 divisions. The rank ablation in Figure~\ref{fig:sens_anal}d shows rapid gains at small ranks, from \(r=0.693\) at rank 25 to \(r=0.813\) at rank 100, followed by a clear plateau, with \(r\) remaining between 0.831 and 0.836 for ranks 150--250. Since these ranks correspond to only a small parameter ratio relative to the full space--time--source data tensor used for merging, this plateau indicates that TMerge achieves near-saturated performance with a parsimonious low-rank representation. Taken together, the sensitivity results show that TMerge remains effective under sparse reference coverage while improving with denser observations, performs best with CONUS-wide three-month fitting blocks despite modest sensitivity to temporal and spatial partitioning, and reaches a performance plateau with a parsimonious low-rank parameterization.

Overall, the CPC-resolution merging experiments show that low-rank tensor structure can be used for satellite--reference precipitation fusion over extended spatiotemporal domains. TMerge substantially improves IMERG estimates across space, time, and precipitation-intensity regimes, demonstrating that low-rank spatiotemporal structure is not only an intrinsic property of precipitation fields but also a practical mechanism that can be exploited to improve satellite precipitation products. 
The sensitivity analyses further show that this improvement is strongest when precipitation is represented over continent-wide, multi-month tensor blocks, indicating that regional long-term spatiotemporal organization provides useful information for merging.

\subsubsection{High resolution experiment}

\begin{table}
\caption{Performance comparison of TMerge, MLP, Linear, QM, and IMERG for daily precipitation estimation at IMERG native resolution across CONUS during 2019--2022. Values are reported as mean $\pm$ standard deviation across runs. The best method per metric is shown in bold.}
\centering
\renewcommand{\arraystretch}{1.25}
\setlength{\tabcolsep}{4pt}
\begin{tabular*}{\textwidth}{@{\extracolsep{\fill}}lccc}
\toprule
Method & Correlation Coefficient ($r$) & RMSE & MAE \\
\midrule
TMerge & \textbf{0.854 $\pm$ 0.002} & \textbf{3.272 $\pm$ 0.026} & \textbf{1.638 $\pm$ 0.018} \\
MLP & 0.771 $\pm$ 0.008 & 4.012 $\pm$ 0.095 & 1.698 $\pm$ 0.050 \\
Linear & 0.534 $\pm$ 0.002 & 6.093 $\pm$ 0.058 & 2.235 $\pm$ 0.033 \\
QM & 0.558 $\pm$ 0.002 & 5.848 $\pm$ 0.055 & 2.191 $\pm$ 0.034 \\
IMERG & 0.534 $\pm$ 0.002 & 6.320 $\pm$ 0.061 & 2.317 $\pm$ 0.035 \\
\bottomrule
\end{tabular*}

\label{tab:total_method_comparison_high_res}
\end{table}

\begin{figure}[]
    \centering
    \includegraphics[width=\linewidth]{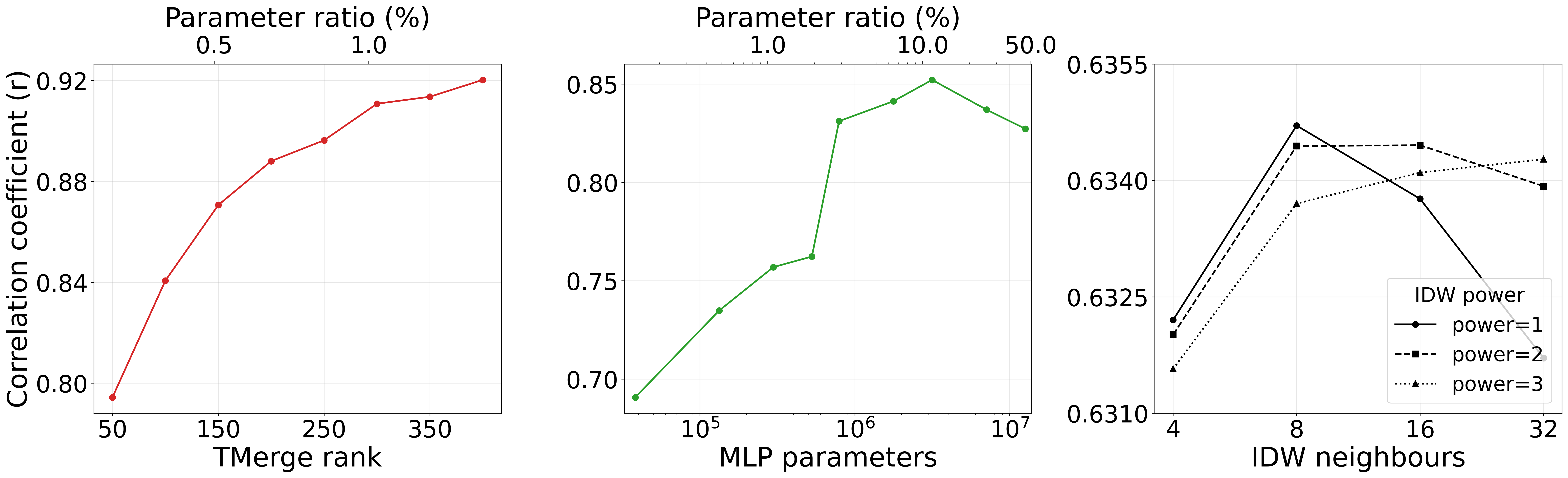}
    \caption{Sensitivity of TMerge performance to rank, MLP performance to architecture size, and QM performance to the number of IDW neighbors and the IDW power, evaluated using the 2018 period.}
    \label{fig:hyperparams_abl}
\end{figure}

The CPC-resolution experiment above provides a controlled evaluation at matched spatial resolution, and this section complements it by testing TMerge using sparse CPC reference constraints to correct IMERG while retaining its finer 0.1° spatial resolution. In each run, 90\% of CPC locations, corresponding to approximately 14\% of native IMERG grid cells, were used as sparse reference constraints, and the remaining 10\% were withheld for testing. As in the CPC-resolution experiment, all models were fitted independently over non-overlapping three-month periods. The sensitivity of each correction method to its hyperparameters was evaluated using the 2018 data, as shown in Figure~\ref{fig:hyperparams_abl}. The best-performing MLP architecture and QM configuration identified in 2018 were retained for subsequent testing, while a rank of 350 was selected for TMerge because further increases produced only marginal gains. Accordingly, the 2018 data were used exclusively for hyperparameter selection, and all results reported below are based on testing in 2019--2022.

\begin{figure}
    \centering
    \includegraphics[width=\textwidth]{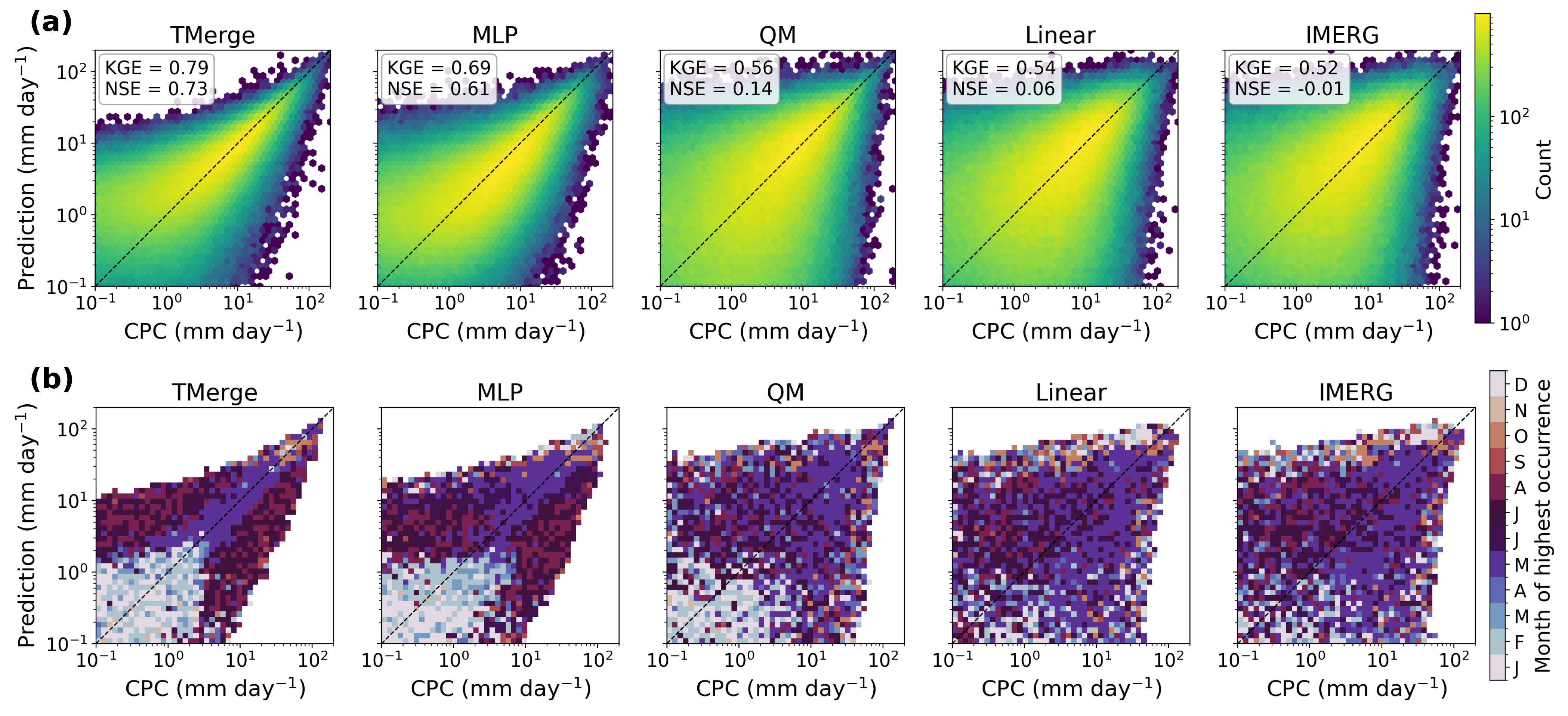}
    \caption{Predicted versus reference (CPC) daily precipitation at withheld CONUS test cells (2019–2022) for TMerge, MLP, QM, Linear, and IMERG: (a) density scatter colored by point count with KGE and NSE annotated and (b) bivariate density colored by modal month, both on log axes with the dashed 1:1 line.}
    \label{fig:conus_bivar_density}
\end{figure}

As reported in Table~\ref{tab:total_method_comparison_high_res}, TMerge achieved the best performance at the native IMERG spatial resolution across CONUS during 2019--2022. IMERG obtained \(r=0.534\), RMSE \(=6.320\) mm day\(^{-1}\), and MAE \(=2.317\) mm day\(^{-1}\). Linear correction and QM provided only modest improvements, reducing RMSE by 3.6\% and 7.5\%, respectively. The MLP baseline performed substantially better, increasing the correlation to \(r=0.771\) and reducing RMSE by 36.5\% relative to IMERG. TMerge nevertheless achieved the strongest improvement, increasing the correlation to \(r=0.854\) while reducing RMSE by 48.2\% and MAE by 29.3\%, reaching RMSE \(=3.272\) mm day\(^{-1}\) and MAE \(=1.638\) mm day\(^{-1}\). Despite using roughly one-tenth of the relative parameter budget of the MLP, TMerge achieved higher accuracy across all metrics, reducing RMSE by a further 18.4\% while increasing correlation. The small standard deviations across runs indicate that these gains are robust to the selection of sparse CPC reference constraints. Overall, these results show that TMerge can incorporate sparse reference information while preserving the finer spatial resolution of IMERG, providing more accurate high-resolution precipitation estimates than the uncorrected IMERG and correction baselines.

Figure~\ref{fig:conus_bivar_density} presents density scatter and bivariate density plots of predicted versus reference (CPC) daily precipitation at the withheld CONUS test cells during 2019--2022, with the 1:1 line denoting perfect agreement. As seen in Figure~\ref{fig:conus_bivar_density}a TMerge exhibits the tightest concentration of points along the 1:1 line and the highest skill (KGE $=0.79$, NSE $=0.73$), followed by MLP (KGE $=0.69$, NSE $=0.61$). QM and Linear predictions remain more diffusely scattered around the line and provide only marginal skill (NSE $=0.14$ and $0.06$, respectively), while raw IMERG is effectively unskilled (NSE $=-0.01$). All methods demonstrate a common conditional bias, with the point cloud lying above the 1:1 line at low intensities and bending below it at high intensities, indicating overestimation of light precipitation and underestimation of heavy rainfall. This departure is smallest for TMerge, whose distribution follows the 1:1 line most closely into the moderate-to-heavy intensity range, showing that it reproduces daily precipitation better than other correction methods or the uncorrected IMERG, particularly at higher intensities. In Figure~\ref{fig:conus_bivar_density}b each bin is colored by its modal month, defined as the month in which precipitation of that magnitude occurs most frequently, revealing the seasonal structure of the prediction--reference relationship. For TMerge, and to a lesser extent MLP, the modal-month pattern forms a smooth gradient along the 1:1 line, with cool-season months concentrated at lower intensities and warm-season months at moderate-to-high intensities. In contrast, QM, Linear, and IMERG show more mottled and mixed seasonal patterns, particularly away from the 1:1 line. Overall, the density plots show that TMerge follows the seasonal structure of CPC daily precipitation more closely than IMERG, with predictions more tightly concentrated about the 1:1 line.

\begin{figure}[]
    \centering
    \includegraphics[width=\textwidth]{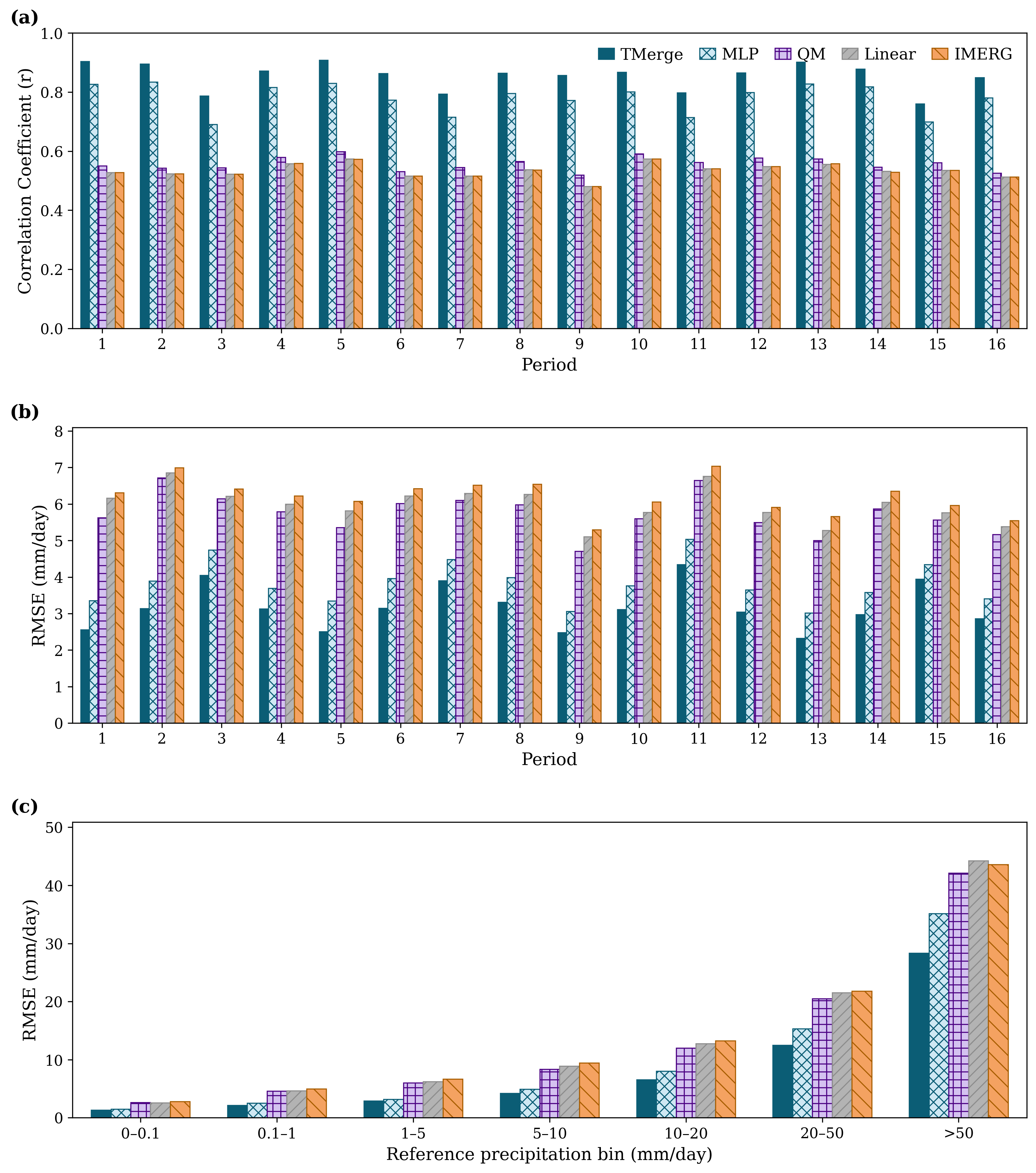}
    \caption{Performance comparison of TMerge, MLP, QM, Linear correction, and the IMERG Final product against the CPC reference at IMERG resolution over CONUS during 2019--2022. (a) Correlation coefficient ($r$) across three-month periods. (b) RMSE across three-month periods. (c) RMSE across CPC reference precipitation bins.}
    \label{fig:ires_seasonal_bins_merge}
\end{figure}

\begin{figure}
    \centering
    \includegraphics[width=0.7\linewidth]{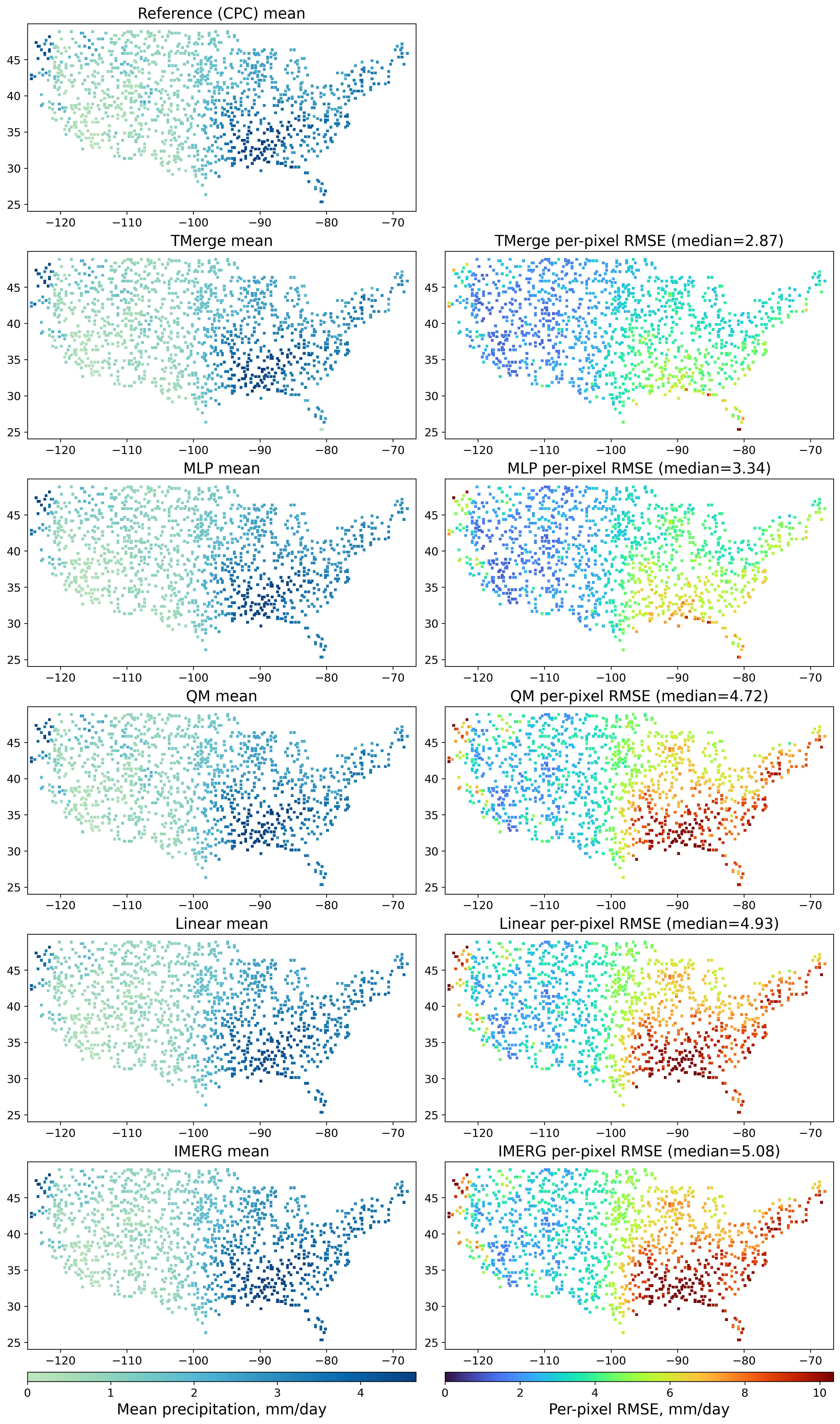}
    \caption{Temporally averaged precipitation and pixel-wise RMSE at the withheld CONUS cells during 2019--2022. For visibility, each test cell is shown as an enlarged marker covering approximately \(115\times\) its true 0.1\(^{\circ}\) grid-cell area. All remaining CPC locations within CONUS were used for model fitting, while areas outside the CONUS domain were excluded from both fitting and evaluation.}
    \label{fig:ires_compare_agg}
\end{figure}

Figure~\ref{fig:ires_seasonal_bins_merge} shows that TMerge achieved the highest correlation and lowest RMSE in every three-month period during 2019--2022, followed consistently by MLP. Averaged across the 16 periods, TMerge reduced RMSE by 48.8\% relative to IMERG, compared with 38.3\% for MLP, while QM and Linear provided only modest improvements; all methods also showed lower accuracy in summer than in winter. Across precipitation-intensity bins, TMerge maintained the lowest RMSE in every regime, reducing IMERG RMSE by approximately 52--56\% for light-to-moderate precipitation and by 34.9\% even for the rare \(>50\) mm day\(^{-1}\) events. This behavior is consistent with the CPC-resolution experiment, showing that TMerge remains effective across seasons and precipitation intensities while preserving the native IMERG resolution. Figure~\ref{fig:ires_compare_agg} compares the temporally averaged precipitation fields and the pixel-wise RMSE, computed across the daily time series at each of the withheld CONUS cells during 2019--2022. TMerge most closely reproduced the CPC spatial pattern, with near-zero spatial-mean bias (\(-0.02\) mm day\(^{-1}\)), and achieved the lowest median per-pixel RMSE (2.87 mm day\(^{-1}\)), compared with 3.34 for MLP and 5.08 for IMERG. TMerge reduced RMSE relative to IMERG at 99.5\% of the withheld cells, with a mean reduction of 46.2\%, demonstrating spatially consistent improvement across the domain.

\begin{figure}[H]
    \centering    
    \includegraphics[width=\textwidth]{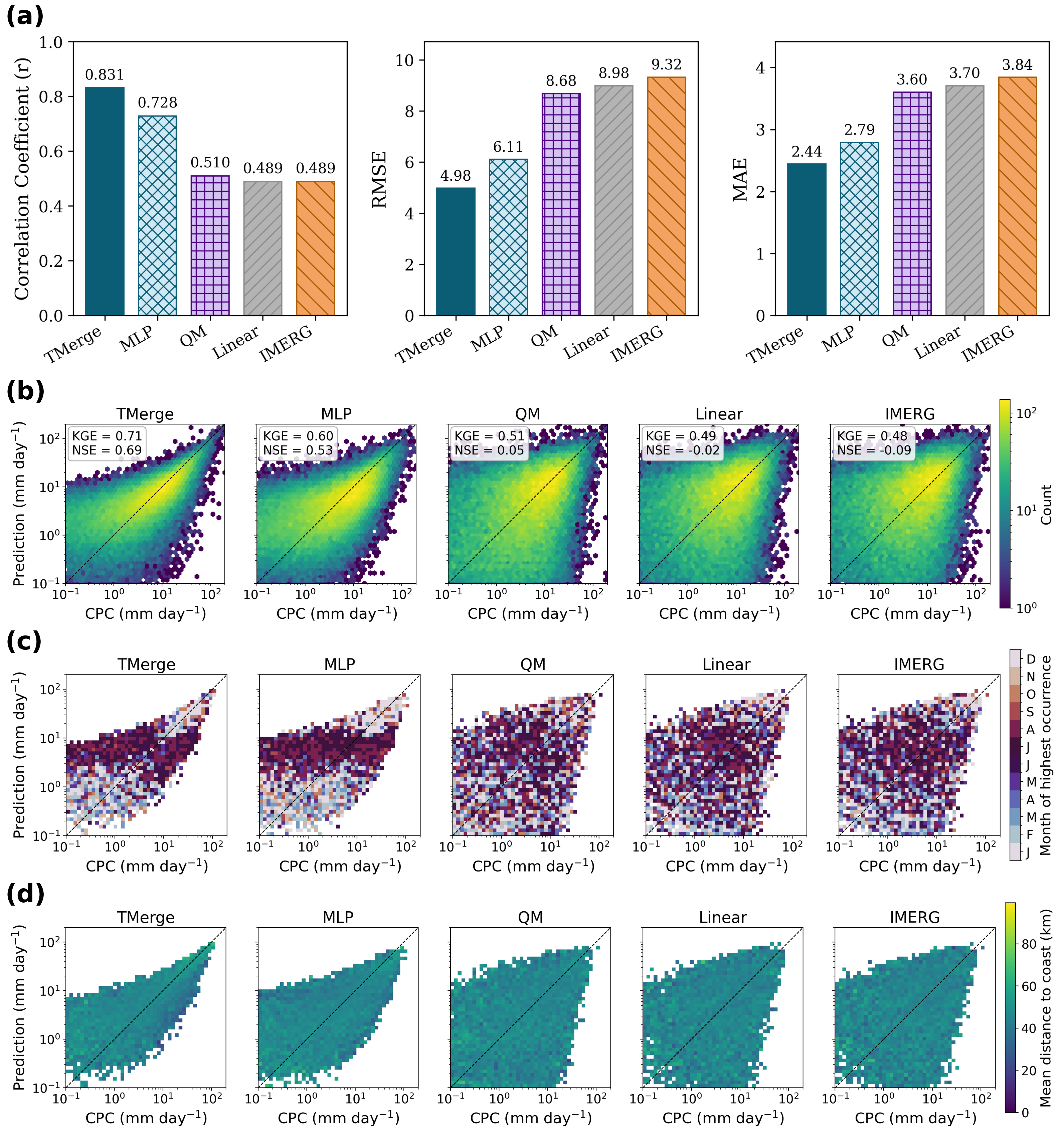}
    \caption{Performance comparison of TMerge, MLP, QM, Linear correction, and the IMERG Final product against the CPC reference at native IMERG spatial resolution over the CONUS coastal zone during 2019--2022: (a) correlation coefficient (\(r\)), RMSE, and MAE, (b) density scatter colored by point count with KGE and NSE annotated, (c) bivariate density colored by modal month, and (d) bivariate density colored by mean distance to the coastline.}
    \label{fig:coastal}
\end{figure}

Given the importance of coastal precipitation and the elevated IMERG errors observed along the West, Gulf, and East Coasts of the United States, we further evaluated performance at withheld test cells located within 100~km of the coastline (Figure~\ref{fig:coastal}). The coastal zone was delineated using the Natural Earth 50~m ocean polygons~\citep{patterson_naturalearth} by computing the distance from each IMERG grid cell to the nearest ocean cell and retaining cells within 100~km. TMerge achieved the strongest coastal performance, increasing the correlation from \(r=0.489\) for IMERG to \(r=0.831\), while reducing RMSE from 9.32 to 4.98~mm~day\(^{-1}\) and MAE from 3.84 to 2.44~mm~day\(^{-1}\), corresponding to reductions of 46.5\% and 36.4\%, respectively (Figure~\ref{fig:coastal}a). TMerge also substantially outperformed the MLP, which achieved \(r=0.728\) and RMSE \(=6.11\)~mm~day\(^{-1}\). The density plots similarly show that TMerge predictions remain most closely concentrated around the 1:1 line, with the highest KGE and NSE (Figure~\ref{fig:coastal}b), while preserving a clearer seasonal organization of precipitation intensities than the other methods (Figure~\ref{fig:coastal}c). Coloring the distributions by mean distance to the coast reveals no strong or sharply separated distance-dependent pattern, although in TMerge and MLP the lighter-colored bins (representing locations farther from the coastline) appear slightly more concentrated near the 1:1 line, whereas QM, Linear, and IMERG show a more intermixed color distribution (Figure~\ref{fig:coastal}d). Overall, the density plots show that TMerge reduced the overestimation of low precipitation values and, more substantially, the underestimation at moderate-to-high intensities by shifting the distribution closer to the 1:1 line, with the remaining underestimation slightly more pronounced at locations that are nearer the coastline.

The resulting TMerge parameterization represented approximately 1.2\% of the space--time--source tensor, whereas the MLP contained 3.15 million trainable parameters, equivalent to approximately 11.5\% of the tensor size, for each 3-month period. Despite using roughly one-tenth of the MLP's relative parameter budget, TMerge achieved higher accuracy across all metrics, as discussed above. Taken together, the results demonstrate that TMerge improves IMERG consistently across seasons, precipitation intensities, and locations, while preserving its native 0.1$^\circ$ resolution with a parsimonious parameterization.

\section{Discussion}

By preserving the native spatial and temporal organization of daily precipitation, tensor factorization captures its dominant variability with greater accuracy and parsimony than matrix-based factorization. This property is important for hydrology because it enables high-dimensional precipitation fields to be represented compactly without disrupting their underlying spatiotemporal structure. This has important implications for a broad range of hydrologic and Earth system applications in which low-rank structure supports efficient data representation, reliable reconstruction, scalable modeling, and systematic monitoring. TMerge exemplifies one such application allowing IMERG and CPC to share low-rank spatial and temporal factors while retaining source-specific information.

TMerge's superior performance over simpler linear and distribution matching methods, as well as the more complex nonlinear MLP, shows that the multilinear structure imposed by tensor factorization can capture rich spatiotemporal relationships with a compact parameterization. This compactness reduces reliance on the large reference datasets typically required by data-intensive neural networks with millions of parameters and can improve generalization to unseen locations. The contrast between the two resolution settings further highlights the parameter efficiency of TMerge relative to the MLP. The MLP performed substantially better applying the native IMERG resolution than using the coarser CPC resolution, where fewer training samples constrained the useful network capacity. TMerge, however, remained effective with a compact parameterization in both settings, indicating that the low-rank formulation is less affected by spatial resolution as long as the dominant spatiotemporal structure can be captured. These characteristics make TMerge particularly well suited for improving satellite precipitation estimates in regions with limited reference-observation coverage.

The stronger performance obtained with CONUS-wide spatial coverage and three-month periods suggests that low-rank modeling is more effective when precipitation is represented as a single tensor over an extended spatiotemporal domain. Although IMERG provides products at subdaily resolutions, this work used the daily IMERG precipitation product because it matches the temporal resolution of the CPC gauge-based reference, is directly relevant to many hydrologic applications, and reduces the computational demands of multi-year, CONUS-wide experiments conducted in this study. Nevertheless, this work's findings motivate extending the analysis to subdaily precipitation, where similar low-rank structure may also be present. Future studies can also extend TMerge to incorporate auxiliary variables, such as topography, as additional tensor modes.

\section{Conclusion}
This study demonstrates that daily precipitation contains an exploitable low-rank structure that is represented more accurately and parsimoniously when its native spatial and temporal dimensions are preserved in tensor form. Building on this property, TMerge integrates dense satellite estimates with sparse reference observations through shared low-rank spatial and temporal factors, substantially improving IMERG precipitation estimates while retaining their native spatial resolution. Across CONUS, TMerge consistently outperformed linear correction, quantile mapping, and a nonlinear MLP across seasons, precipitation-intensity regimes, and levels of reference availability, with particularly strong improvements in coastal regions where satellite errors were greatest. These findings establish low-rank tensor structure as an organizing property of precipitation variability with potential benefits for several hydrologic and Earth system applications, and demonstrate its effectiveness in improving high-resolution precipitation estimates over extended regions with limited reference-observation coverage.

\section*{Data Statement}
The Integrated Multi-satellitE Retrievals for GPM (IMERG) Final Run daily precipitation product (GPM\_3IMERGDF, Version~07) used in this study is available from NASA GES DISC at \url{https://doi.org/10.5067/GPM/IMERGDF/DAY/07}~\citep{Huffman2020IMERG}. The Climate Prediction Center (CPC) Unified Gauge-Based Analysis of Daily Precipitation over CONUS is available from NOAA PSL at \url{https://downloads.psl.noaa.gov/Datasets/cpc_us_precip/} \citep{Chen2008CPC}. The Python code for this study is available at \url{https://github.com/rmsolgi/TMerge.git} under the MIT License. Predictions and data underlying the figures are available at Zenodo via \url{https://zenodo.org/records/21745127} under a Creative Commons Attribution 4.0 International (CC BY 4.0) license \citep{solgi2026sup}.

\bibliographystyle{elsarticle-harv} 
\bibliography{mybib}
\end{document}